%% file: main.tex
\documentclass[conference]{IEEEtran}
\IEEEoverridecommandlockouts
\usepackage{cite}
\usepackage{amsmath,amssymb,amsfonts}
\usepackage{algorithmic}
\usepackage{graphicx}
\usepackage{textcomp}
\usepackage{graphics} % for pdf, bitmapped graphics files
\usepackage{epsfig} % for postscript graphics files
\usepackage{mathptmx} % assumes new font selection scheme installed
\usepackage{times} % assumes new font selection scheme installed
\usepackage{amsmath} % assumes amsmath package installed
\usepackage{amssymb}  % assumes amsmath package installed
 \usepackage[table]{xcolor}
\usepackage{hyperref}
\usepackage{cleveref}
\usepackage{booktabs}
\usepackage{tabularx}
\usepackage{float}
\def\BibTeX{{\rm B\kern-.05em{\sc i\kern-.025em b}\kern-.08em
    T\kern-.1667em\lower.7ex\hbox{E}\kern-.125emX}}
    
\IEEEoverridecommandlockouts
\begin{document}

\title{Forecasting the Success of Television Series using Machine Learning}

\author{\IEEEauthorblockN{1\textsuperscript{st} Ramya Akula}
\IEEEauthorblockA{\textit{Industrial Engineering} \\
\textit{University of Central Florida}\\
Orlando, USA \\
ramya.akula@knights.ucf.edu}
\and
\IEEEauthorblockN{1\textsuperscript{st} Zachary Wieselthier}
\IEEEauthorblockA{\textit{Industrial Engineering} \\
\textit{University of Central Florida}\\
Orlando, USA \\
zwieselthier@knights.ucf.edu}
\and
\IEEEauthorblockN{2\textsuperscript{nd} Laura Martin}
\IEEEauthorblockA{\textit{Industrial Engineering} \\
\textit{University of Central Florida}\\
Orlando, USA \\
lauraconrad@knights.ucf.edu}
\and
\IEEEauthorblockN{2\textsuperscript{nd} Ivan Garibay}
\IEEEauthorblockA{\textit{Industrial Engineering} \\
\textit{University of Central Florida}\\
Orlando, USA \\
igaribay@ucf.edu}
}

\maketitle
\begin{abstract}
Television is an ever-evolving multi-billion-dollar industry. The success of a television show in an increasingly technological society is a vast multi-variable formula. The art of success is not just something that happens, but is studied, replicated, and applied. Hollywood can be unpredictable regarding success, as many movies and sitcoms that are hyped up and promise to be a hit end up being box office failures and complete disappointments. In current studies, linguistic exploration is being performed on the relationship between Television series and target community of viewers. Having a decision support system that can display sound and predictable results would be needed to build confidence in the investment of a new TV series. The models presented in this study use data to study and determine what makes a sitcom successful. In this paper, we use descriptive and predictive modeling techniques to assess the continuing success of television comedies: \textit{The Office}, \textit{Big Bang Theory}, \textit{Arrested Development}, \textit{Scrubs}, and \textit{South Park}. The factors that are tested for statistical significance on episode ratings are character presence, director, and writer. These statistics show that while characters are indeed crucial to the shows themselves, the creation and direction of the shows pose implication upon the ratings and therefore the success of the shows. We use machine learning based forecasting models such as linear regression, K Nearest Neighbors, Stochastic Gradient Descent, Decision Tree and Forests, Neural Network, and Facebook Prophet, to accurately predict the success of shows. The models represent a baseline to understanding the success of a television show and how producers can increase the success of current television shows or utilize this data in the creation of future shows. Due to the many factors that go into a series, the empirical analysis in this work shows that there is no one-fits-all model to forecast the rating or success of a television show. However, because the variables are statistically significant, they are still able to optimize and affect the rating positively.
\end{abstract}


\input{body.tex}

\section*{Acknowledgement}
We thank Sean Worth and Ryan Nicholls for collecting and preparing the datasets for this work.

\bibliographystyle{IEEEtran}
\bibliography{bibliography}

\end{document}

%% file: body.tex
\section{Introduction} 
According to Variety\footnote{\href{https://variety.com/}{Variety - Source of Entertainment Business News}}, an authoritative and trusted source of entertainment business news, there is an increase in production of many neoteric TV sitcoms each year. The report\footnote{\href{https://variety.com/2016/tv/news/peak-tv-2016-scripted-tv-programs-1201944237/}{TV Peaks Again in 2016: Could It Hit 500 Shows in 2017?}} states that 455 original scripted programs aired on American television in 2016, and throughout a decade the amount of scripted TV will increase by 137\%. With all of these new TV programs being released each year only several of them take hold and become popular with the public. Major companies like Amazon and Netflix have been spending record amounts of money on TV sitcoms; A record\footnote{\href{https://variety.com/2018/digital/news/netflix-700-original-series-2018-1202711940/}{Netflix Eyeing Total of About 700 Original Series in 2018}} blasts on Television industry that Netflix is set to spend over \$ 8 billion on content in 2018 with around 700 original TV sitcoms. The ratings are a major factor in determining the popularity of a TV sitcom. Linguistic analysis in  \cite{rey2014changing , bubel2005linguistic , signes2007we, baker2006public , paltridge2011genre , wodak2009discourse , richardson2006dark , bednarek2010language} determines a systematic analytical framework and empirical data about the expression of ideology through dialogues in episodic television. According to work conducted in \cite{bednarek2015overview}, TV screen-writing research, the perspectives are distinguished as (1) Creation of television script-writing and kinds of processes involved in the production of different versions of series. (2) The characteristic features of the outcome of television writing, i.e., the nature of scripts or episodes/series broadcast. (3) Engaging the audience with television writing through analysis of discourses produced by fans, surveys, focus groups, interviews, and questionnaires. The review\cite{bednarek2015overview} raises awareness of linguistic research on TV dialogue, and encourages interdisciplinary perspectives and collaborative research projects on television writing, in screenwriting research and beyond. Having a predictive machine learning algorithm that uses at least one complete season of data will tell you whether another season is worth producing. The previous works are heavily dependent on qualitative analysis with hardly one or two case studies in particular. We employ artificial intelligence based machine learning algorithms\footnote{\href{https://github.com/akula01/TV-Shows-Sitcom-Success.git}{Our GitHub Repository: Success Prediction of Television Series using Machine Learning}} to understand the accomplishments through forecast models of five TV series: \textit{The Office},\textit{ The Big Bang Theory}, \textit{Arrested Development}, \textit{Scrubs}, and \textit{South Park}.

Organization of this paper is as follows: Section 2 consists of related work in this field. Section 3 explains the proposed approach along with problem formulation, data description, and preparation of five TV shows. Section 4 describes the results and analysis of forecast models on each TV show. Section 5 briefly concludes the paper and is followed by the future scope of extension.

% \usepackage{booktabs}
\begin{table*}[]
    \centering
    \begin{tabular}{@{}|l|c|c|c|c|c|c|@{}}
    \toprule
    \multicolumn{7}{|c|}{{\color[HTML]{3531FF} \textbf{Descriptive Statistics of Sitcoms}}}                                                                                                                                                                                                                                                              \\ \midrule
    {\color[HTML]{680100} \textbf{Sitcoms}} & {\color[HTML]{680100} \textbf{Episodes}} & {\color[HTML]{680100} \textbf{Seasons}} & {\color[HTML]{680100} \textbf{Mean Episode Rating}} & {\color[HTML]{680100} \textbf{Median Episode Rating}} & {\color[HTML]{680100} \textbf{Normal Distribution}} & {\color[HTML]{680100} \textbf{Shapiro-Wilk Test}} \\ \midrule
    \textbf{The Office}                            & \textbf{186}                             & \textbf{1-9}                            & \textbf{8.2}                                        & \textbf{8.3}                                          & {\color[HTML]{036400} \textbf{Yes}}                 & \textbf{0}                                        \\ \midrule
    \textbf{The Big Bang Theory}                   & \textbf{231}                             & \textbf{1-10}                           & \textbf{8.1}                                        & \textbf{8.1}                                          & {\color[HTML]{CB0000} \textbf{No}}                  & \textbf{0.021}                                    \\ \midrule
    \textbf{Arrested Development}                  & \textbf{66}                              & \textbf{1-5}                            & \textbf{8.4}                                        & \textbf{8.5}                                          & {\color[HTML]{CB0000} \textbf{No}}                  & \textbf{0.001}                                    \\ \midrule
    \textbf{Scrubs}                                & \textbf{117}                             & \textbf{1-5}                            & \textbf{8.3}                                        & \textbf{8.2}                                          & {\color[HTML]{CB0000} \textbf{No}}                  & \textbf{0}                                        \\ \midrule
    \textbf{South Park}                            & \textbf{79}                              & \textbf{1-5}                            & \textbf{8.1}                                        & \textbf{8.2}                                          & {\color[HTML]{CB0000} \textbf{No}}                  & \textbf{0.0027}                                   \\ \bottomrule
    \end{tabular}
    \caption{Descriptive statistics of sitcoms. For each sample of data we calculate mean and median of episodes' ratings, check for normality of these ratings, otherwise we perform Shapiro-Wilk test on abnormal episodes' ratings}
    \label{tab:dataset_stats}
\end{table*}

\begin{figure}[h!]
    %\centering
    \includegraphics[scale=0.5]{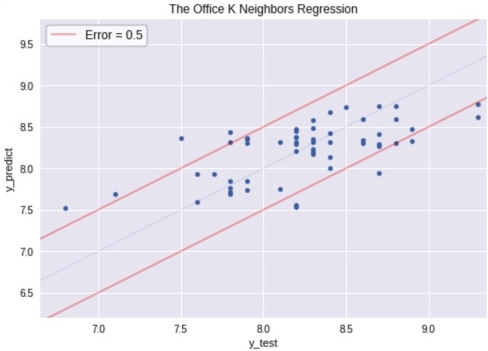}
    \caption{Shows correlation between actual values and predicted values}
    \label{fig:correlation}
\end{figure}

\section{Related Work}
Television shows are the products that exhibit real-world cultural and traditional differences, primarily emphasized through dialogues. Many studies focus on dialogue nature in TV series \cite{quaglio2009television , bednarek2010language , bednarek2011language , tagliamonte2005so , richardson2010television}  and humor in comedy series \cite{stokoe2014dispreferred , brock2011bumcivilian , urios2011prosody , bo2008humorous}. For instance, how \textit{The Office} has adapted to the institutional context, culture, and humor of the United States as mentioned in \cite{beeden2010office}, its success as a British sitcom, illustrates that national identity as a vital part of the global television format trade. While the growth of format adaptations may reflect the increasingly globalized contemporary world format adaptations encourage articulations of national identity and cultural belonging. A variation description in title sequences for American television series is earned\cite{bednarek2014and} by exploring the key features of the television title sequence – such as length, credits, characters, sound and style – and relates these to its functional characteristics. As a result, it provides a synchronic snapshot of a significant contemporary cultural product that viewers regularly engage with and enjoy. This survey not only shows the variation inherent in these cultural products but also pinpoints areas that are worth investigating more closely and in depth. From the product perspective, studies in \cite{maras2011some, macdonald2013screenwriting} show that there is no final script on which to base firm conclusions and pointing to the importance of other texts that refer to the screen idea. However, according to \cite{nannicelli2012ontology} this object problem affects not only TV scripts, but also television works, primarily because of media convergence, fan texts, and serialization. Humor can also be studied pragmatically within the framework of the relevance theory, which proposes that communication\cite{hu2012analysis} is a process of ostension and inference and differentiates the maximal relevance and optimal relevance. The work in \cite{beeden2010office} analyses how television format adaptations work through articulations of national identity and suggest that the success of an adaptation may link to its ability to reflect and interpret its new context. Despite the global success of the sitcom genre, there are clear differences in the situations, characters, and humor used by British and American sitcoms which must be addressed by an adaptation. Based on the analysis of keywords and trigrams, the study conducted in \cite{bednarek2012get} explores characteristics of contemporary American English television dialogue. The results suggest that the expression of emotion is a key defining feature of the language of television, cutting across individual series and different televisual genres. The previous works are heavily dependent on qualitative analysis with one or two case studies in particular. We employ artificial intelligence based machine learning algorithms to understand and predict the accomplishments with forecast models on five TV series. To that end in this area, our work is never attempted before, especially with this number of sitcoms as case-studies.

\section{Proposed Method}
When it comes to sitcoms, the most famous ones stay popular even after their generation. \textit{The Office}, \textit{The Big Bang Theory}, \textit{Arrested Development}, \textit{Scrubs}, and \textit{South Park} have withstood the test of time and remain popular. In this work we use the data from these sitcoms with their corresponding factors that represent the successful components of a sitcom: title, director, writer, original air date, rating, and how many lines each character spoke. 
\begin{figure}[h!]
    %\centering
    \includegraphics[scale=0.8]{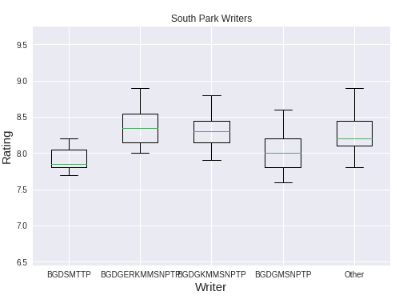}
    \caption{South Park: Box Plot Visualization for rating of an Episode vs the Writer}
    \label{fig:southpark_box}
\end{figure}
\subsection{\textbf{Problem Description}}
\begin{figure*}[h!]
    \centering
    \includegraphics[scale=0.45]{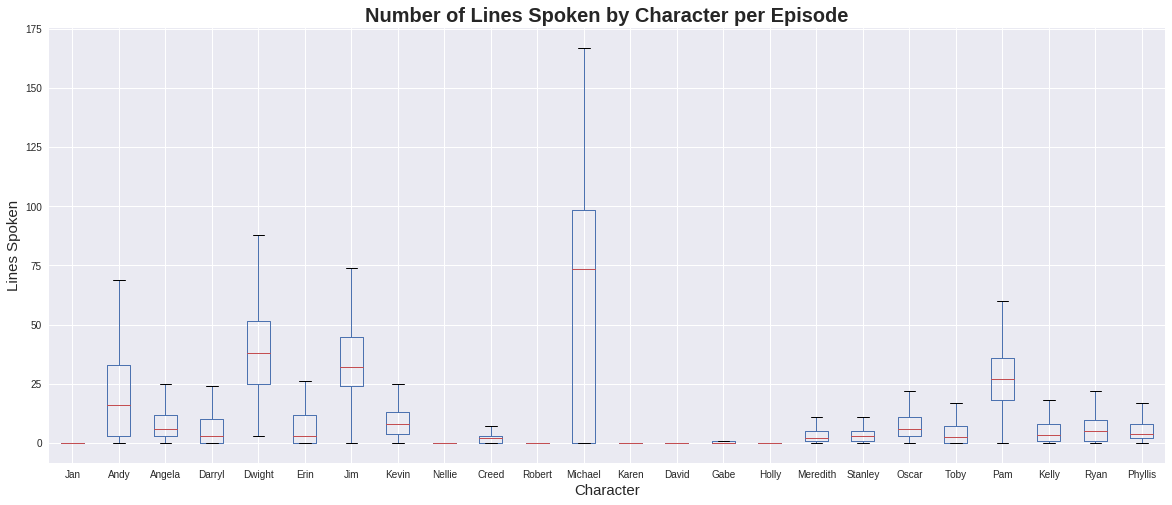}
    \caption{The Office: Box Plot Visualization for number of lines spoken by character in an Episode}
    \label{fig:office_box}
\end{figure*}

\begin{figure*}[h!]
    \centering
    \includegraphics[scale=0.5]{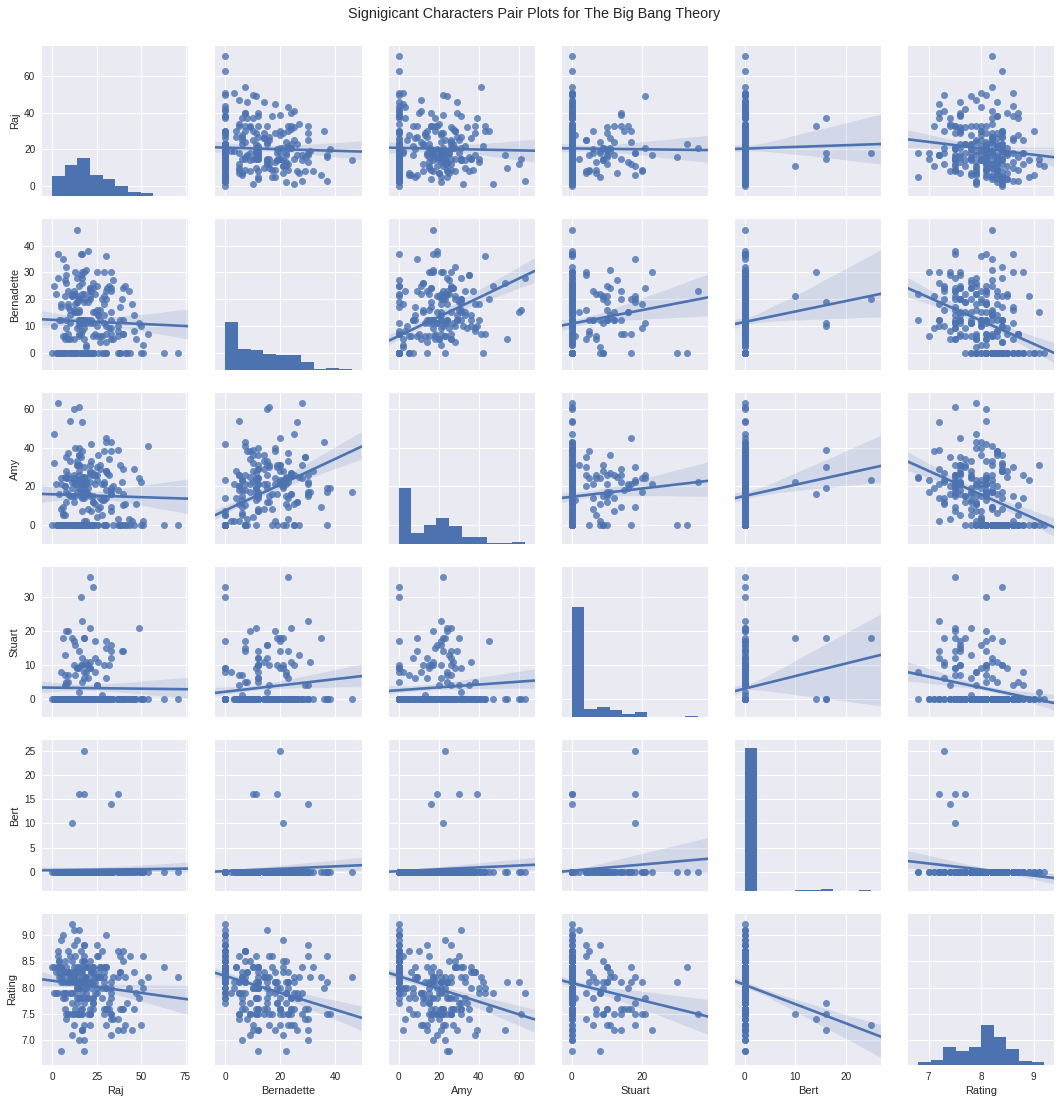}
    \caption{\textit{The Big Bang Theory}: Pair Plot Visualization for significant characters in an Episode}
    \label{fig:bigbang_pairplot}
\end{figure*}

In this work, we use descriptive statistics, visualizations, hypothesis testing, and predictive analytics to understand the differences and impacts that different factors have on our dependent variable, episode rating. This process simplifies the options into two competing hypotheses:
\begin{itemize}
    \item \textbf{\textit{H0:}} The mean or median episode rating  is the same through the tested factors (Director, Writer, and Number of lines spoken by popular characters)
    \item \textbf{\textit{HA:}} The mean or median episode rating differs with the tested factors (Director, Writer, and Number of lines spoken by popular characters)
\end{itemize}
Descriptive statistics is an important step used to understand the data. For instance, if we find that subset of the data does not have a normal distribution, we will not be able to perform an analysis of variance (ANOVA) \cite{girden1992anova}; instead, we use the Kruskal Wallis test \cite{breslow1970generalized} for non-parametric data. This test will tell if there is a significant difference between the medians of two or more groups. When the data is not normal the median will give a better estimate of the center point. Furthermore, the visualizations are useful in identifying trends within our data. Box plots \cite{williamson1989box} help to display the distribution of the data and how different distributions compare as shown in Figure \ref{fig:southpark_box}. Scatter plots \cite{keim2010generalized} show if there is an apparent relationship between two different variables. A sample box plot visualization for \textit{The Office} dataset is shown in Figure \ref{fig:office_box} and a sample pair plot visualization for \textit{The Big Bang Theory} dataset is shown in Figure \ref{fig:bigbang_pairplot}. 

\begin{figure}[h!]
    \centering
    \includegraphics[scale=0.4]{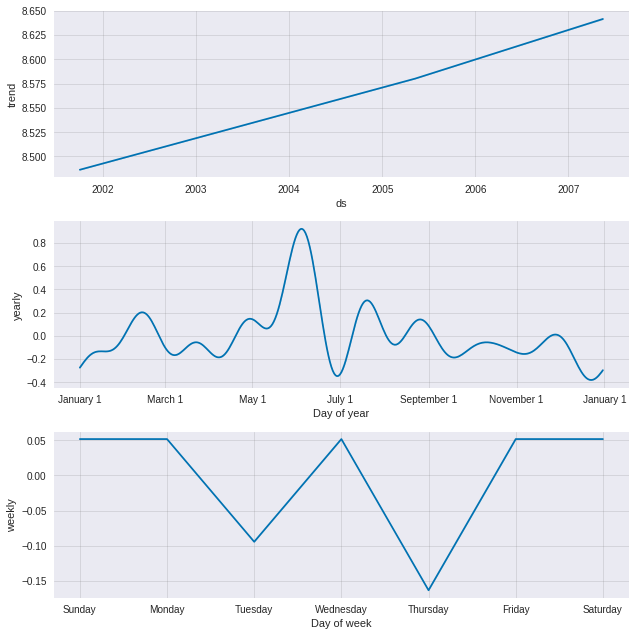}
    \caption{Scrubs: Forecast Regression Model Performance}
    \label{fig:scrubs_forecast}
\end{figure}

\begin{figure*}[h!]
    \centering
    \includegraphics[scale=0.8]{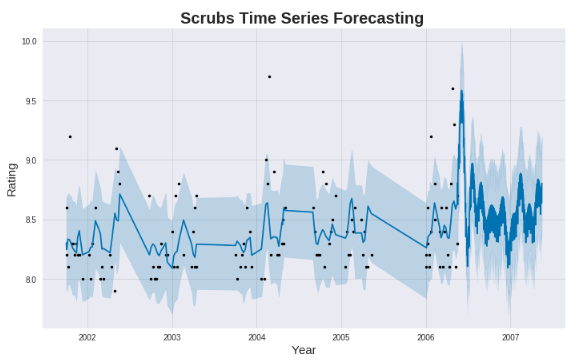}
    \caption{Scrubs: Time Series Forecast using Facebook Prophet}
    \label{fig:scrubs_prophet_forecast}
\end{figure*}

\begin{figure*}[h!]
    \centering
    \includegraphics[scale=0.75]{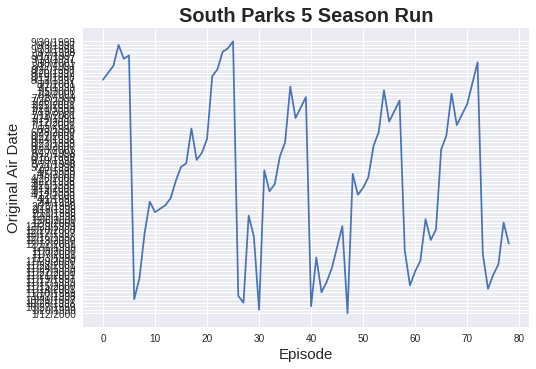}
    \caption{South Park: Seasonal Forecast using Facebook Prophet}
    \label{fig:southpark_seasons}
    \end{figure*}

Hypothesis tests\cite{lehmann2006testing} are used to verify the trends that we have identified statistically. The linear regression model \cite{aalen1989linear} will tell us the correlation between a characters presence and the episode’s rating; the more lines a character speaks the more presence they have in an episode. These models will help identify what factors lead to a successful episode. For predictive analytics, we use time series form of data that measures how things change over time, stored in time order. Prophet\footnote{\href{https://facebook.github.io/prophet/}{Facebook Prophet - Forecasting at Scale}} is a procedure for forecasting time series data based on an additive model where non-linear trends are fit with yearly, weekly, and daily seasonality, plus holiday effects. A sample Prophet seasonal forecast visualization for the \textit{South Park} dataset is shown in Figure \ref{fig:southpark_seasons}. It works best with time series that has strong seasonal effects and several seasons of historical data. Prophet is robust to missing data and shifts in the trend, and typically handles outliers well. As stated before each episode has the corresponding factor of original air date and rating which is used to understand how the episodes perform over time. A sample Prophet forecast visualization for the \textit{Scrubs} dataset is shown in Figure \ref{fig:scrubs_prophet_forecast}. To predict the outcome of future events we use machine learning \cite{nasrabadi2007pattern} based forecast regression techniques such as: Linear Regression\cite{aalen1989linear}, K Nearest Neighbors\cite{cover1967nearest}, Scholastic Gradient Descent\cite{zemel2001gradient}, Decision Tree and Forests \cite{tong2003decision}, Neural Networks \cite{sarle1994neural}. A sample machine learning based regression model's performance for the \textit{Scrubs} dataset is shown in Figure \ref{fig:scrubs_forecast}. To assess the fit of our regression models, we calculate $ R^2 $ (R squared)  and Root Mean Square Error (RMSE) as shown in Table \ref{tab:reg_perform}. Due to the page limit, we cannot include corresponding images that show the performances of other machine learning based regression models, box plot, pair-plot, Prophet seasonal visualizations for remaining datasets; however, all the visualizations are available on our GitHub.

\subsection{\textbf{Data Description}}
For this work, we extrapolate the data on title, directors, writers, original air date, and ratings from Internet Movie Database (IMDb)\footnote{\href{https://www.imdb.com/}{Internet Movie Database (IMDb)}} and the sitcom transcripts are collected using a web-scrapper with the sources that contain an episodes' transcripts and then searches the HTML for a matching character and counts the character's number of lines.  We collect corresponding episodes', writers, and directors information from Wikipedia. We clean the data by removing insignificant characters that will skew the data; that is, we remove the characters that spoke less than five lines per episode and spoke less than five times throughout the show. Table  \ref{tab:dataset_stats} shows the descriptive statistics of sitcoms: \textit{The Office}, \textit{The Big Bang Theory}, \textit{Arrested Development}, \textit{Scrubs}, and \textit{South Park}. For each episodes' rating, we calculate the mean, median, and Shapiro-Wilk to check for normality. Additionally, histograms further check normality.

\section{Results}
The impact of character presence on the episode's rating is calculated by the correlation of the number of lines spoken by the character to the episode's rating. The box plot shows the distribution of lines spoken by each reoccurring character throughout the series’ run. The linear regressions find, that a majority of the cast's number of lines spoken per episode does not have a statistically significant impact. However, the combination of all the characters in the model does. ANOVA is used to test the difference between directors and writers on episode rating. Multiple machine learning algorithms use the inputs of lines spoken by characters, director, and writer to predict an episodes rating. Table \ref{tab:reg_perform} shows the performance of machine learning based regression models on these datasets.

\subsection{\textbf{The Office:}} The supporting cast, the majority of characters, is defined by 25 lines spoken or less per episode. The main cast, Michael, Pam, Jim, and Dwight can be separated by speaking more than 25 lines. Michael has the highest average number of lines, but also the largest distribution, with 25\% of episodes having no lines spoken by Micheal. All characters have outliers, suggesting some episodes focus more on certain characters than others. The regression’s  $ R^2 $ is 0.4, suggesting the model explains the variance in episodes' well. The characters Nellie and Robert have a p-value less than .05, signifying their statistical significance. Both also hurt episode rating with a coefficient of -0.0137 and -0.0122 respectively. The ANOVA for writer and director are statistically significant, meaning that there is a difference of episode rating between the factors. In the test, director Matt Sohn resulted individually statistically significant, with a coefficient value of -0.6 and an average rating of 7.8. The QQ plot is mostly normal with an increase in scattering near the outliers. The statistically significant writers are Brent Forrester and Greg Daniels. Their corresponding coefficients are 0.550 and 0.650 and an average rating of 8.56 and 8.66,  respectively, indicating a positive impact on episode ratings. Visual inspection of the time series plot shows some random variability over the eight years, but the time series also seems to have a decreasing downward trend in the last two seasons, which aired in 2012 and 2013. The trend may be attributed to Nellie and Robert who appear in the last to seasons and hurt the show’s rating. The trend for the rating time series appears to be linear and stationary over time. \textit{The Office’s} historical air dates show a repeating pattern over successive periods indicating a recurring seasonal pattern. The K Nearest Neighbors approach, which uses similarly matched known data to predict future data, performed the most successfully. The model can predict a majority of the tested episodes accurately within half a unit rating and around half of the tested episodes within a quarter of a rating, as shown in Figure \ref{fig:correlation}. Furthermore, the model can predict all tested episodes within a one-unit rating. The model can explain 43\% of the variance between episodes' ratings. Other factors can include acting performance, written line quality, and plot.

 \begin{figure}[h!]
    %\centering
    \includegraphics[scale=0.5]{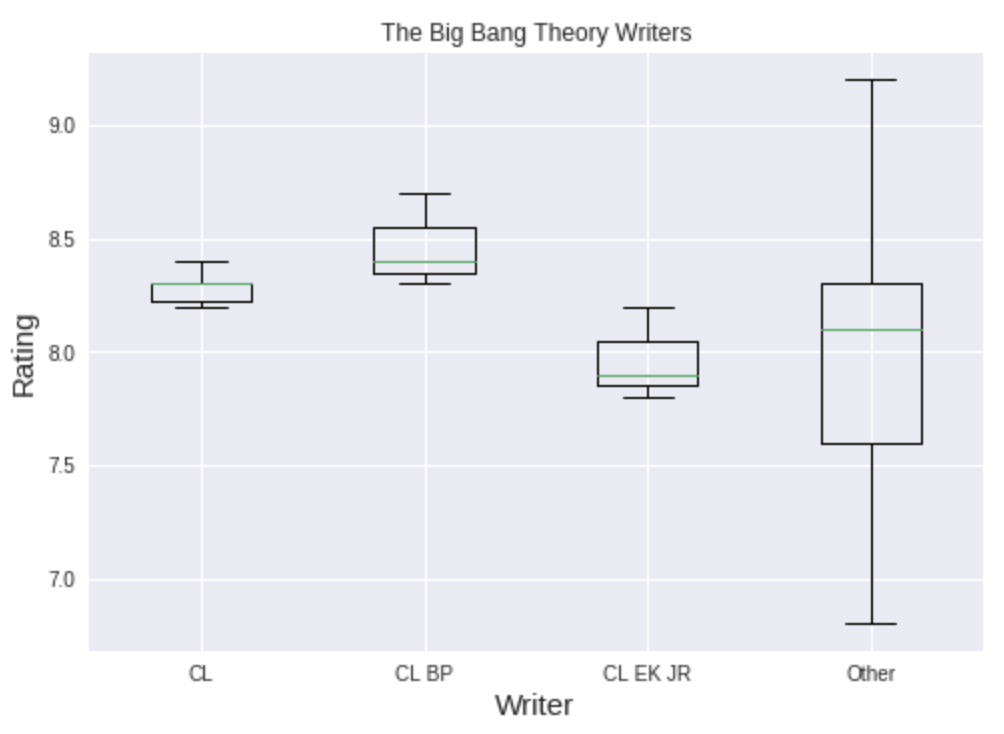}
    \caption{The Big Bang Theory: Writer Combination Analysis}
    \label{fig:writers}
    \end{figure}
    
\subsection{\textbf{The Big Bang Theory:}} 
\begin{figure}[h!]
    %\centering
    \includegraphics[scale=0.55]{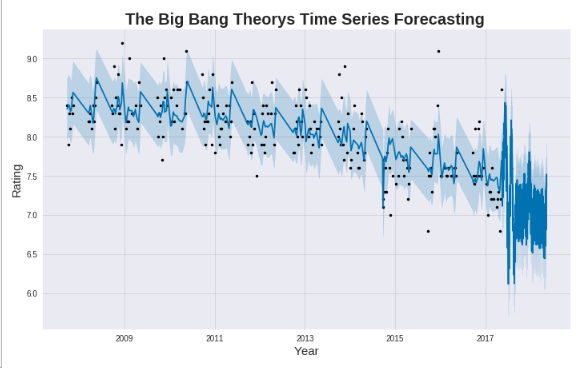}
    \caption{The Big Bang Theory: Noticeable Downward Trend}
    \label{fig:time_big}
    \end{figure}

The five main characters, Leonard, Sheldon, Penny, Howard, and Raj spoke most of the lines and have a varying median ranging between 18 to 50 lines. The linear regression results with the majority of individual characters not having a statistically significant correlation on the episode's rating. However, the combination of all the characters do result statistically significant. The model suggests that character presence has an impact on the rating of an episode. The model can explain 29\% of the variance between episodes. The statistically significant characters are Raj, Bernadette, Amy, Stuart and Bert; with P values of 0.023, 0.001, 0.003, 0.011, and 0.021, respectively. Moreover, the main characters all had corresponding negative coefficients of -0.005, -0.011, -0.007, -0.010, and -0.023 respectively. Evidence that their presence may have a negative impact on episode rating. Writer and director impact is tested using Shapiro-Wilk to account for the data set’s non-normal distribution. There is not enough evidence to suggest a difference between directors; however there is enough evidence for a statistical difference between writers: p = 0.26, 0.01. The box plot shown in Figure \ref{fig:writers} below visualizes this difference, the combination "Chuck Lorre and Bill Prady” performed the best of the writers and “Chuck Lorre, Eric Kaplan, and Jim Reynolds” performed the worst. The "Other" group writers who wrote less than five episodes, have the largest variance with  maximum and minimum rating. \textit{The Big Bang Theory} time series exhibits a downward linear trend pattern as shown in Figure \ref{fig:time_big}. Visual inspection of the time series plot shows some random variability over the show’s 5-year history. The trend for the rating time series appears to be linear and stationary over time. \textit{The Big Bang Theory}’s historical air dates show a repeating pattern over successive periods indicating a seasonal pattern. Every season which consists of approximately 20 episodes shows a recurring seasonal pattern. Each season which consists of approximately 20 episodes is a recurring pattern. The K Nearest Neighbors approach, which uses similarly matched known data to predict future data, performed the most successfully. The K Nearest Neighbors algorithm with the three inputs can explain 17\% of the variance between episodes, and the mean squared error is 0.13.
    
\subsection{\textbf{Arrested Development:}}  
\begin{figure}[h!]
    %\centering
    \includegraphics[scale=0.8]{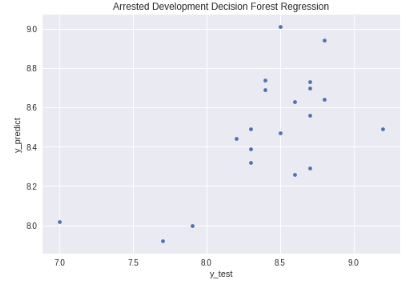}
    \caption{Arrested Development: Decision Forest Regression Model}
    \label{fig:ad_dforest}
    \end{figure}

Michael spoke the largest average of lines in each episode. The linear regression results with the majority of individual characters not having a statistically significant correlation on the episode's rating. However, the combination of all the characters is statistically significant, suggesting that character presence has an impact on the rating of an episode. The model’s $ R^2 $ of 0.606, which means that  61\% of the variance can be well explained. The character Lucille is statistically significant with a p-value of 0.024. Moreover, Lucille has a corresponding negative coefficient of -0.006. This is evidence that her presence has a negative effect on episode rating; however, the corresponding scatter plot suggests that an outlier may cause this. Writer and director impact is tested using Shapiro-Wilk to account for the data’s non-normal distribution. There is no evidence to reject the null hypotheses for writers.  There is evidence of a significant difference for directors. The combinations of "Mitchell Hurwitz" and "Troy Miller" performed significantly worse. The "Other" group, directors who directed less than five episodes, has the second largest variance and the maximum rating. The \textit{Arrested Development} time series exhibits an upward linear trend pattern from 2003 to 2006. All episodes aired at the same time when the show was brought back in 2013; this causes the time series model to be inaccurate for predicting future ratings. Visual inspection of the time series plot shows some random variability over the show’s 3-year history. The trend for the rating time series appears to be linear and stationary over time. The decision forest approach, which builds multiple decision trees and merges them to get a more accurate and stable prediction, performed the most successfully as shown in Figure \ref{fig:ad_dforest}. The decision forest algorithm can explain 27\% of the variance between episodes, and the mean squared error is 0.152. 
    
\subsection{\textbf{Scrubs:}}  The main characters, Turk, Dr.Cox, J.D., Elliot, and Carla, spoke more than 20 lines per episode and have the largest distributions. The regression is not statistically significant with a p-value of 0.218, suggesting that character presence does not have an impact on the rating of an episode. The linear regression results in the majority of individual characters' correlation on the episode's rating not statistically significant. The character Elliot is statistically significant with a p-value of 0.041 and a corresponding coefficient of -0.007. This is evidence that her presence has a negative effect on episode rating. Writer and director impact is tested using Shapiro-Wilk to account for the data’s non-normal distribution. Both the writer and director hypothesis test fail to reject the null hypothesis, suggesting these two factor do not have a significant impact on the shows rating. The \textit{Scrubs} time series exhibits a slight upward linear trend pattern from 2001 to 2007, with ratings increasing from 8.2 to 8.5. Visual inspection of the time series plot shows some random variability over the six years. The trend for the rating time series does not appear to be stationary over time. The Scrub’s historical air dates do not show a repeating pattern over successive periods indicating there is not a seasonal pattern. The time series forecasts a horizontal linear trend line with an average rating of 8.5. None of the tested algorithms were able to explain the variance between episodes well. However, the decision forest, which builds multiple decision trees and merges them to get a more accurate and stable prediction, performed the most successfully. The decision forest algorithm can explain 12\% of the variance between episodes, and the mean squared error is 0.096.
    
\subsection{\textbf{South Park:}}The three main characters, Cartman, Kyle, and Stan spoke the majority of the lines and had a median ranging between 30 to 60. Kyle and Stan have similar distributions, while Cartman has a much wider distribution. The regression for number of lines spoken by characters is not statistically significant. The p-value of 0.121, suggests that character presence does not have an impact on the rating of an episode. The linear regression results in the majority of individual characters not having a statistically significant correlation on the episode's rating. While the model is not statistically significant, the character Terrance does have a statistically significant impact with a p-value of 0.001. His negative coefficient value of -0.0157 shows that the presence of Terrance has a negative correlation with the episode's rating. Writer and director impact is tested using Shapiro-Wilk to account for the data’s non-normal distribution. There is no evidence to reject the null hypotheses for directors.  There is evidence of a significant difference for writers, p-value 0.01. The box plot shown in Figure \ref{fig:southpark_box} visualizes the difference in episode ratings; “Brian Graden, David R. Goodman, Erica Rivinoja, Kyle McCulloch, Matt Stone, Nancy M. Pimentel, and Trey Parker” (designated as BGDGERKMMSNPTP) slightly performed the best of the writers and “Brian Graden, Dan Sterling, Matt Stone, Trey Parker” (designated as BGDSMSTP) performed the worst. The \textit{South Park} time series exhibits a slight upward linear trend pattern from 1997 to 2001, with ratings increasing from 8.0 to 8.5. Visual inspection of the time series plot shows some random variability over four years. The trend for the rating time series does not appear to be stationary over time. \textit{South Park}’s historical air dates do not show a repeating pattern over successive periods, indicating there is not a seasonal pattern. None of the tested algorithms were able to explain the variance between episodes.

% \usepackage{booktabs}
\begin{table*}[]
    \begin{tabular}{@{}lcccccccccc@{}}
    \toprule
    \multicolumn{1}{c}{{\color[HTML]{000000} \textbf{Sitcoms}}} & \multicolumn{2}{c}{{\color[HTML]{680100} \textbf{The Office}}}                & \multicolumn{2}{c}{{\color[HTML]{680100} \textbf{The Big Bang Theory}}}       & \multicolumn{2}{c}{{\color[HTML]{680100} \textbf{Arrested Development}}}     & \multicolumn{2}{c}{{\color[HTML]{680100} \textbf{Scrubs}}}                     & \multicolumn{2}{c}{{\color[HTML]{680100} \textbf{South Park}}}              \\ \midrule
    {\color[HTML]{000000} \textbf{Regression Model Evaluation}}        & {\color[HTML]{00009B} \textbf{$ R^2 $}}    & {\color[HTML]{00009B} \textbf{RMS}}   & {\color[HTML]{00009B} \textbf{$ R^2 $}}    & {\color[HTML]{00009B} \textbf{RMS}}   & {\color[HTML]{00009B} \textbf{$ R^2 $}}   & {\color[HTML]{00009B} \textbf{RMS}}   & {\color[HTML]{00009B} \textbf{$ R^2 $}}    & {\color[HTML]{00009B} \textbf{RMS}}    & {\color[HTML]{00009B} \textbf{$ R^2 $}}    & {\color[HTML]{00009B} \textbf{RMS}} \\
    Linear                                                             & -0.505                                & 0.337                                 & -0.14                                 & 0.185                                 & -4.16                                & 1.082                                 & -1.85                                 & 0.323                                  & -1.16                                 & 0.531                               \\
    K Nearest Neighbors                                                        & {\color[HTML]{036400} \textbf{0.398}} & {\color[HTML]{036400} \textbf{0.135}} & {\color[HTML]{036400} \textbf{0.176}} & {\color[HTML]{036400} \textbf{0.134}} & -0.17                                & 0.245                                 & 0.043                                 & 0.105                                  & -0.53                                 & 0.37                                \\
    Stochastic Gradient Descent                                        & -24.145                               & 5.636                                 & -7.221                                & 1.339                                 & -32.28                               & 6.966                                 & -10.234                               & 1.24                                   & -18.71                                & 4.8                                 \\
    Decision Tree                                                      & 0.147                                 & 0.191                                 & 0.17                                  & 0.134                                 & -0.271                               & 0.266                                 & -0.274                                & 0.147                                  & -1.75                                 & 0.67                                \\
    Neural Network                                                     & 0.321                                 & 0.15                                  & -0.051                                & 0.171                                 & -0.415                               & 0.296                                 & -0.248                                & 0.138                                  & {\color[HTML]{036400} \textbf{-0.24}} & {\color[HTML]{036400} \textbf{0.3}} \\
    Decision Forest                                                    & 0.33                                  & 0.15                                  & 0.173                                 & 0.135                                 & {\color[HTML]{036400} \textbf{0.40}} & {\color[HTML]{036400} \textbf{0.126}} & {\color[HTML]{036400} \textbf{0.163}} & {\color[HTML]{036400} \textbf{0.0921}} & -0.4                                  & 0.34                                \\ \bottomrule
    \end{tabular}
    \caption{Performance of Machine Learning based Regression Models on Sitcom Datasets}
    \label{tab:reg_perform}
\end{table*}

\section{Conclusion}
While the production of television is an art, we suggest that it can also be a science. Having a decision support system that predicts success can build confidence in the investment of a new TV series. The models presented used historical data to study and determine what makes a sitcom great. While not all series are successful, the factors: character presence, director, and writer are shown to be accurate predictors of a show's IMBD ratings. Furthermore, the writer and director have a stronger impact than character presence. Besides, while also not always successful, machine learning algorithms were able to forecast episode's rating more accurately. \textit{The Office} performed the most successful in forecasting, with an ability to predict all tested episodes within one unit of the actual rating and an {$ R^2 $} of 0.398. In contrast to \textit{The Office}'s success, South Park was unable to be successfully forecasted. This work shows that there is not a one-size-fits-all model for predicting rating or success, possibly due to the immense amount of factors that go into the production of a television show. The models presented represent a baseline for understanding the success of a television show and how producers can increase the success of current television shows or utilize data in the creation of future shows. Further analysis needs to focus on a plethora of other factors, including demographic, advertising budget, internet presence, theme, and schedule to fully explain the entirety of the variance of a show’s rating. However, because the variables character presence, director, and writer were statistically significant, they are still able to be optimized to improve the rating. 

